%% file: main.tex
\documentclass[runningheads]{llncs}
\usepackage{eccv}
\usepackage{hyperref}

\usepackage{graphicx}
\usepackage{booktabs}
\usepackage[accsupp]{axessibility}

\usepackage{multirow}
\usepackage{makecell}
\usepackage{amsmath}

\usepackage{orcidlink}
\usepackage{pdfpages}
\usepackage[utf8]{inputenc}
\usepackage[T1]{fontenc}
\usepackage{CJKutf8}
\usepackage{threeparttable}
\usepackage{xcolor}

\usepackage{booktabs}      
\usepackage{multirow}      
\usepackage{graphicx}      
\usepackage[table]{xcolor} 
\usepackage{threeparttable} 

\DeclareUnicodeCharacter{3131}{\CJKchar{mj}{`ㄱ}}

\newcommand{\figref}[1]{Fig.~\ref{#1}}
\newcommand{\tabref}[1]{Tab.~\ref{#1}}

\newcommand{\figureref}[1]{Figure~\ref{#1}}

\begin{document}

\title{RainODE: Continuous-Time Precipitation Forecasting with Latent Neural ODEs} 

\titlerunning{RainODE}

\author{Yeeun Seong\textsuperscript{*}\orcidlink{0009-0008-8389-1019}\and
Doyi Kim\textsuperscript{*}\orcidlink{0000-0002-9849-3127}\and
Minseok Seo\orcidlink{0000-0002-5940-120X} \and
Changick Kim\textsuperscript{$\dagger$}\orcidlink{0000-0001-9323-8488}}

\authorrunning{Y.~Seong et al.}

\institute{
Korea Advanced Institute of Science and Technology (KAIST), Daejeon, South Korea\\
\email{\{yeeunseong,doyi.kim,minseok.seo,changick\}@kaist.ac.kr}
}

\maketitle

{\let\thefootnote\relax\footnotetext{
\textsuperscript{*}Equal contribution.\\
\textsuperscript{$\dagger$}Corresponding author.
}}

\begin{abstract}
In precipitation forecasting, not only accuracy but also temporal resolution is critical.
However, increasing temporal resolution is constrained by observational limitations and the computational cost of dense discrete modeling.
To overcome this limitation, we reformulate precipitation forecasting as a continuous-time dynamical system and propose RainODE, a framework that models precipitation evolution in latent space using a Neural ODE.
This formulation enables derivative-consistent temporal dynamics and captures the dominant large-scale advective motion of precipitation systems.
Nevertheless, a purely deterministic ODE struggles to represent non-advective intensity changes such as localized growth, decay, and sub-grid variability, often leading to over-smoothed predictions.
To address this issue, we introduce a stochastic source modeling module based on a Brownian Bridge formulation, which refines residual intensity variations and restores fine-grained structures while preserving advective consistency.
By combining deterministic continuous dynamics with stochastic refinement, RainODE enables arbitrary-time inference while maintaining sharp predictions.
Experiments on SEVIR and the newly introduced Radar-based Precipitation Integrated Dataset (RAPID) demonstrate consistent improvements across multiple temporal intervals and precipitation regimes.
The code is available at \url{https://github.com/SeongYE/RainODE}.
\end{abstract}

\section{Introduction}
\label{sec:intro}
Data-driven short-term precipitation forecasting has rapidly advanced with the emergence of large-scale benchmark datasets and diverse deep learning architectures~\cite{shi2015convolutional, an2025deep, gao2022earthformer,gao2023prediff, yu2024diffcast, gong2024cascast, yoon2024probabilistic, lin2025alphapre,hoonextreme, sun2026stormdit}.
In particular, the SEVIR~\cite{veillette2020sevir} dataset has become a widely adopted benchmark, where predicting 12 future timestamps at 5-minute intervals, corresponding to one hour ahead, serves as a standard evaluation protocol.
Despite strong performance under such discrete settings, precipitation in the Earth system evolves continuously over time.
The atmosphere does not change in isolated 5-minute increments; rather, intensity, structure, and motion evolve as a coupled dynamical process~\cite{wen2021effect, otsuka2016precipitation}.

However, practical forecasting models are inevitably trained on discretely sampled snapshots due to observational and modeling constraints.
This mismatch between continuous physical processes and discrete prediction frameworks introduces structural limitations.
While discrete models can achieve high accuracy and temporal consistency at predefined intervals, such consistency is typically tied to the specific training grid.

\begin{figure}[t!]
  \centering
  \includegraphics[width=\linewidth]{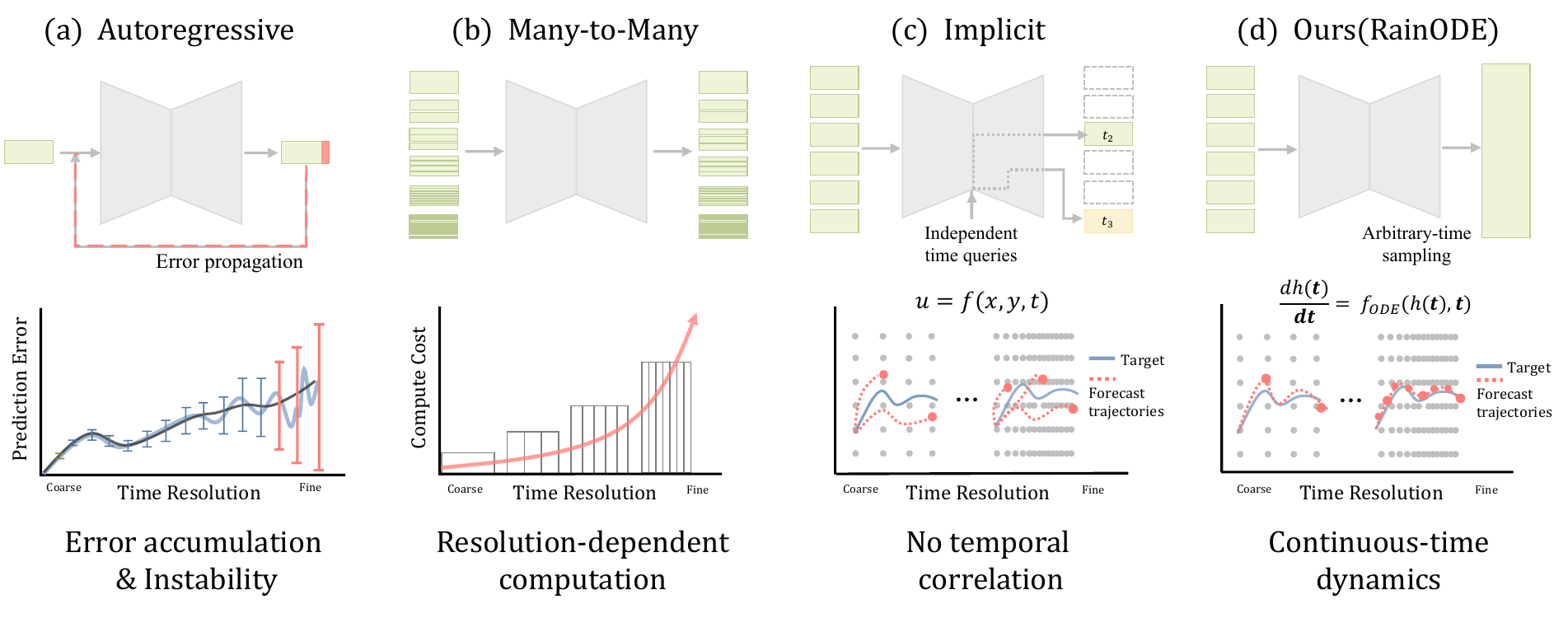}
  \caption{\textbf{Task Description.} Increasing temporal resolution causes (a) error accumulation, (b) computational bottlenecks, and (c) temporally uncorrelated predictions. In contrast, \textbf{(d) RainODE} models shared continuous-time latent dynamics via a Neural ODE, enabling stable long-horizon forecasting and arbitrary-time inference.}
  \label{fig:fig1}
\end{figure}

Existing forecasting paradigms have to densify discrete predictions to address higher temporal resolution.
As shown in~\figref{fig:fig1}, autoregressive models suffer from error accumulation as rollouts become denser~\cite{shi2015convolutional, wang2017predrnn, wang2018predrnn++, wang2018eidetic,wang2019memory, guen2020disentangling,yu2020efficient}, whereas many-to-many models jointly predict dense spatio-temporal blocks at the cost of rapidly increasing computational complexity~\cite{gao2022simvp,tan2025simvpv2,ning2023mimo,tang2024video}.
Implicit models~\cite{sonderby2020metnet, espeholt2022deep, andrychowicz2023deep, kim2024masked} provide flexibility in representing arbitrary time points, but they do not explicitly enforce dynamical consistency across time.
Instead of merely increasing output frequency within fixed temporal grids, we reformulate precipitation forecasting as a continuous-time dynamical process.

We propose RainODE, a framework for continuous-time precipitation forecasting with Latent Neural ODEs. 
RainODE models temporal evolution in latent space through a Neural ODE, learning derivative-consistent dynamics rather than relying on discrete frame-to-frame mappings.
A complementary Stochastic Source Modeling (SSM) module further captures fine-scale spatial variability while preserving large-scale dynamical consistency.
Furthermore, to evaluate generalization beyond trained temporal intervals and to assess long-horizon forecasting performance up to six hours, we introduce RAPID, a new benchmark dataset supporting +6 hour prediction.
Experimental results demonstrate that RainODE not only achieves state-of-the-art performance on both SEVIR and RAPID, but also enables time-axis continuous forecasting across varying temporal intervals.

\section{Related Works}
\subsection{Data-Driven Approaches}
Data-driven precipitation forecasting is commonly formulated as a spatio-temporal sequence modeling problem.
Early autoregressive models~\cite{shi2015convolutional, shi2017deep, wang2017predrnn, zhao2024new} preserve temporal dependency by recursively feeding back predictions, but inevitably accumulate errors over time.
To mitigate this issue, many-to-many (M2M) frameworks~\cite{chen2023swinrdm, sonderby2020metnet, gao2022simvp, bai2022rainformer, gao2022earthformer, lin2025alphapre, hoonextreme, sarabia2025rainpro, feng2025perceptually, yu2024diffcast, ribeiro2025flowcast} jointly predict the entire target horizon in a single forward pass, effectively capturing correlations within a fixed temporal window.
However, extending prediction horizons typically increases computational and memory costs.
Implicit forecasting models take the desired lead time as an input query and directly predict the corresponding future state, enabling flexible lead-time forecasting~\cite{sonderby2020metnet, espeholt2022deep, andrychowicz2023deep, kim2024masked}.
While this design avoids recursive rollout, predictions at different timestamps are typically generated independently, and consistency across outputs is not explicitly modeled.
Nevertheless, most existing approaches are designed and evaluated on fixed discrete temporal intervals, with both training and inference tied to predefined time steps.

\subsection{Neural ODEs in Forecasting}
Neural ODEs~\cite{chen2018neural} were originally proposed to integrate ordinary differential equations into deep neural networks by modeling hidden state dynamics in continuous-time.
Since then, they have been widely adopted in fluid dynamics and dynamical system modeling, where continuous-time formulations are naturally aligned with physical processes.
In meteorology, recent works such as ClimODE~\cite{vermaclimode} and WeatherODE~\cite{liu2024mitigating} apply Neural ODEs to large-scale atmospheric forecasting.
In this context, Neural ODEs are attractive because their continuous-time formulation can, in principle, support temporally flexible forecasting beyond fixed discrete transitions.
These approaches demonstrate strong performance in global, spatially closed system settings.
However, quantitative analysis under increased temporal resolution, particularly across unseen temporal intervals, has not been explicitly investigated.
Moreover, these methods have not been extended to open system, high-resolution precipitation forecasting settings.

\subsection{Temporal Downscaling}
Temporal downscaling methods aim to reconstruct intermediate frames between discretely sampled observations.
Motion-based extrapolation and deep learning interpolation approaches~\cite{tatsubori2022deep, demiray2023efficienttempnet, wang2025temporal} increase temporal density by learning mappings between adjacent timestamps.
In practice, these methods generate intermediate states conditioned on observed frames or outputs from a separate forecasting model within predefined temporal gaps.
As a result, temporal refinement is inherently conditioned on discrete predictions, and overall performance remains strongly dependent on the underlying forecasting model.
In contrast, our approach models precipitation evolution itself as a continuous-time dynamical process, intrinsically extending the temporal axis during prediction.
This enables time-axis continuous forecasting that is not tied to specific discrete intervals, while promoting spatio-temporal consistency and improving predictive performance.

\begin{figure}[tb]
  \centering
  \includegraphics[width=\linewidth]{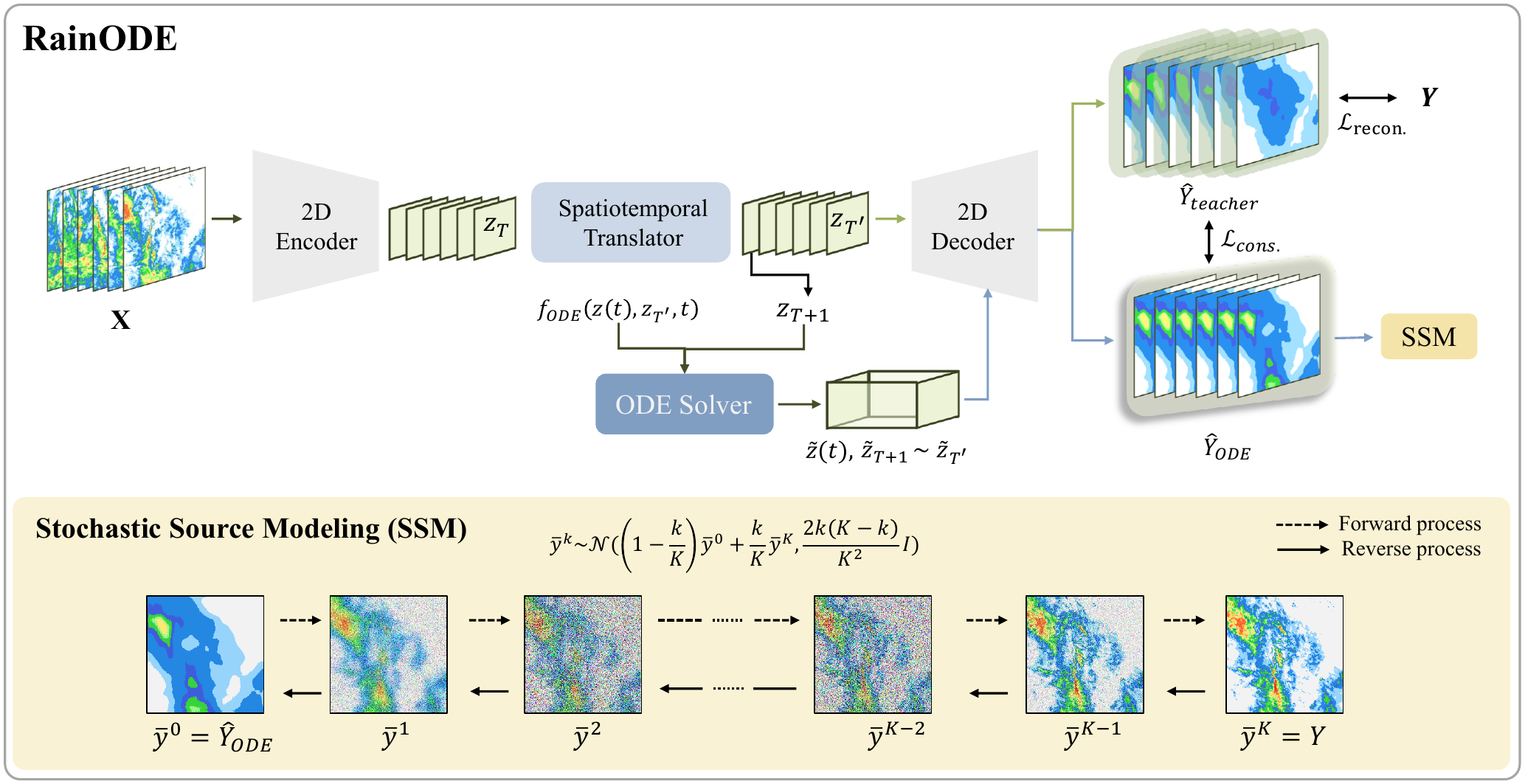}
  \caption{\textbf{Overview of RainODE.} Input radar sequences are encoded into latent features, which evolve continuously via a Neural ODE conditioned on the target forecasting interval. Consistency loss aligns ODE forecasts $\hat{Y}_{\text{ODE}}$ with teacher predictions $\hat{Y}_{\text{teacher}}$, while reconstruction loss supervises $\hat{Y}_{\text{teacher}}$. The stochastic source modeling (SSM) module further refines predictions using a Brownian Bridge diffusion process.}
  \label{fig:main}
\end{figure}

\section{RainODE}
\subsection{Discrete-Time Forecasting Perspective}
Radar-based precipitation forecasting aims to predict future precipitation states $Y \in \mathbb{R}^{T' \times H \times W}$ given past radar observation sequences $X \in \mathbb{R}^{T \times H \times W}$, where $T'$ is the prediction horizon, $H$ and $W$ are the spatial height and width, and $T$ is the number of input frames.

Although existing approaches such as autoregressive, many-to-many, implicit models, and our method differ in their architectural designs, they can be abstracted under a unified learning objective:
\begin{equation}
\min_{\theta} \; \mathcal{L}\big(f_{\theta}(X), Y\big),
\end{equation}
where $f_{\theta}$ denotes a precipitation forecasting model parameterized by $\theta$, and $\mathcal{L}(\cdot)$ measures the discrepancy between predictions and ground truth.

Prior studies mainly improve forecasting performance from two directions: enhancing the model architecture $f_{\theta}$, such as SimVP~\cite{gao2022simvp}, MetaVP~\cite{tan2025simvpv2}, and Earthformer~\cite{gao2022earthformer}, or designing task-specific loss functions $\mathcal{L}$~\cite{yan2024fourier, gao2023prediff}. 
MSE reduces numerical error but often produces overly smooth outputs~\cite{gao2022simvp,gao2022earthformer}, whereas diffusion-based losses yield sharper fields at the expense of higher pixel-wise errors~\cite{gao2023prediff,yu2024diffcast, gong2024cascast}. 
Frequency-domain losses such as Fourier Correlation Loss (FCL) emphasize high-frequency components but may introduce intensity biases~\cite{yan2024fourier}.

Although these approaches improve accuracy under discrete temporal supervision, they do not change the underlying discrete-time formulation. 
\subsection{Continuous-Time Reformulation}
The spatiotemporal evolution of precipitation can be described by a differential equation of the form:
\begin{equation}
\frac{\partial u}{\partial t} = \underbrace{-\mathbf{v} \cdot \nabla u}_{\text{Advection}} + \underbrace{D \nabla^2 u}_{\text{Diffusion / Mixing}} + \underbrace{S(u, t)}_{\text{Source / Sink}}
\label{eq:advection_diffusion_annotated}
\end{equation}

This formulation decomposes precipitation dynamics into three components: 
large-scale advective transport, diffusion or mixing effects, and local source or sink processes such as growth and decay.
Here, $u$, $\mathbf{v}$, $D$, and $S(u,t)$ denote precipitation intensity, horizontal transport velocity, diffusion/mixing, and local growth or decay, respectively.
Guided by Eq.~\eqref{eq:advection_diffusion_annotated}, we focus on the coherent transport-like behavior suggested by the advection term.
This transport-like component often evolves smoothly over time, making it compatible with a continuous-time latent dynamics formulation.
We therefore use a Neural ODE to parameterize latent temporal evolution in precipitation dynamics.

However, precipitation evolution also involves unresolved processes such as turbulence, local growth, and decay, related to the diffusion and source/sink terms in Eq.~\eqref{eq:advection_diffusion_annotated}. 
These effects introduce stochastic and multi-modal intensity variations that cannot be captured by deterministic latent dynamics alone, often leading to over-smoothed predictions.
To account for this uncertainty, we introduce a Brownian Bridge~\cite{li2023bbdm, liu20232}--based conditional refinement that probabilistically models the residual evolution between deterministic forecasts and observed states, restoring high-frequency structures and localized variability (\figref{fig:main}).

\noindent\textbf{Latent ODE Dynamics.} 
We model the deterministic motion component via a Neural ODE in latent space.  Given discrete input frames $X_{0:T}$, a 2D spatial encoder $e(\cdot)$ extracts latent representations $Z_{0:T} = \{z_0, \dots, z_T\}$.
A Translator network $f_{\text{trans}}$ predicts future discrete latent states $Z_{T+1:T'}$.

The ODE is initialized with the first translated latent state: $z(0) = z_{T+1}$.
The continuous latent trajectory is parameterized over a normalized temporal variable $s \in [0,S]$, which represents the forecasting interval. 
The evolution of the latent state is governed by

\begin{equation}
\frac{dz(s)}{ds} = f_{\text{ODE}}(z(s), z(S), s),
\label{eq:ode_derivative}
\end{equation}

where $s$ denotes the continuous-time variable.
The trajectory is obtained by numerically integrating the ODE using a fourth-order Runge--Kutta (RK4)~\cite{runge1895numerische} solver, where $\tau$ represents the continuous integration variable used by the solver:

\begin{equation}
\tilde{z}(s)
=
z(0)
+
\int_{0}^{s}
f_{\text{ODE}}(z(\tau), z(S), \tau)
\, d\tau.
\label{eq:ode_integral}
\end{equation}

This formulation enables temporally arbitrary predictions while enforcing derivative-consistent evolution. At inference, querying the trajectory at arbitrary $s$ enables dense forecasts without retraining, and the decoded prediction is

\[
\hat{y} = d(\tilde{z}(s)).
\]

\noindent\textbf{Consistency-Guided Training.}
To enforce trajectory-level consistency and stabilize training, we adopt a dual-objective training strategy.

During training, both the discrete latent sequence predicted by the Translator and the continuous latent trajectory obtained by integrating the ODE are projected back to the observation space through the shared 2D spatial decoder $d(\cdot)$.
This generates two distinct spatial predictions: the discrete teacher prediction $\hat{Y}_{teacher} = d(Z_{T+1:T'})$ and the continuous ODE prediction $\hat{Y}_{\text{ODE}} = d(\{\tilde{z}(s(t))\}_{t=T+1}^{T'})$.

The total training objective $\mathcal{L}_{\text{Stage I}}$ is defined by two MSE terms. The first component is a task reconstruction loss that directly minimizes the discrepancy between the Translator's discrete output and the actual ground truth sequence $Y$. The second component acts as a consistency regularization, enforcing the continuous predictions generated by the ODE solver to align smoothly with the discrete trajectory. The total loss is mathematically formulated as:

\begin{equation}
\mathcal{L}_{\text{Stage I}} = \underbrace{\frac{1}{N} \sum || \hat{Y}_{teacher} - Y ||_2^2}_{\text{Reconstruction Loss}} + \alpha\underbrace{\frac{1}{N} \sum || \hat{Y}_{\text{ODE}} - \hat{Y}_{teacher} ||_2^2}_{\text{Consistency Loss}}
\label{eq:stage1_loss}
\end{equation}

\noindent\textbf{Stochastic Source Modeling.}
While Stage I enforces derivative-consistent temporal evolution, its deterministic formulation produces over-smoothed predictions and fails to recover fine-grained variability.
To model unresolved components associated with diffusion and source/sink effects in Eq.~\eqref{eq:advection_diffusion_annotated}, we introduce stochastic source modeling (SSM) based on a Brownian Bridge diffusion process.

Let the ODE prediction be $\bar{y}^0 = \hat{Y}_{\text{ODE}}$ and the ground truth be $\bar{y}^K = Y$. 
The forward diffusion process is defined as:

\begin{equation}
\bar{y}^k \sim 
\mathcal{N}
\left(
\left(1 - \frac{k}{K}\right)\bar{y}^0 
+ 
\frac{k}{K}\bar{y}^K,
\;
\frac{2k(K-k)}{K^2} I
\right).
\label{eq:brownian_forward}
\end{equation}
Except for this endpoint-conditioned formulation, we follow the implementation details of \cite{li2023bbdm}.
The mean trajectory gradually bridges the deterministic prediction toward the target state, while the time-dependent variance injects controlled stochastic perturbations.
This provides a practical approximation of unresolved precipitation variability without explicitly solving the governing PDE.

Through this stochastic refinement, Stage II restores high-frequency spatial structures and localized variability while preserving the large-scale advective consistency learned in Stage I.

\section{RAPID Dataset}
\label{sec:dataset}

\subsection{Dataset Construction}
Precipitation nowcasting research has widely adopted the SEVIR dataset~\cite{veillette2020sevir}, which provides storm-centered multimodal observations under a fixed temporal configuration.
To complement existing benchmarks and enable evaluation across multiple temporal scales, we introduce RAPID (Radar-based Precipitation Integrated Dataset), a radar-only benchmark constructed from operational Hybrid Surface Precipitation (HSP) products over the Korean Peninsula\footnote{Those could be found in \url{https://apihub.kma.go.kr}}.

RAPID spans 2022--2025 at 5-minute temporal and 0.5 km spatial resolution. 
For computational consistency, data are cropped to a fixed domain and resampled to 2 km resolution.
Candidate frames are selected if (1) precipitation exceeding 0.1 mm/h occupies more than 10\% of the spatial domain, or (2) the 99th percentile intensity exceeds 10 mm/h. 
This preserves diverse precipitation structures while excluding non-precipitating scenes. 
Forecasting sequences are constructed using a sliding window with six input and six target frames.

\begin{table}[t]
  \centering
  \begin{threeparttable}
    \caption{\textbf{Precipitation intensity statistics} in physical units. Both datasets exhibit highly imbalanced, heavy-tailed distributions, with RAPID showing a wider dynamic range at high intensities.}
  \label{tab:dataset_stats}
  \setlength{\tabcolsep}{5pt}
  \begin{tabular}{l|c|ccccccc}
    \toprule
    Dataset & Res. & Zero & 50th & 90th & 99th & Non-zero 99th & 99.9th & Max  \\
    \midrule
    SEVIR(kg/m$^2$) & 1 km & 44\% & 0.00 & 0.84 & 6.38 & 13.11 & 26.93  & 80  \\
    RAPID(mm/hr) & 2 km & 72\% & 0.00 & 1.15 & 8.88 & 21.28 & 39.17 & 180  \\
    \bottomrule
  \end{tabular}
  \begin{tablenotes}\scriptsize
  \item *SEVIR values are converted to physical units using the decoding formulation provided in the original paper~\cite{veillette2020sevir}.
  \end{tablenotes}
  \end{threeparttable}
\end{table}

\begin{figure}[tb]
  \centering
  \includegraphics[width=1.0\linewidth]{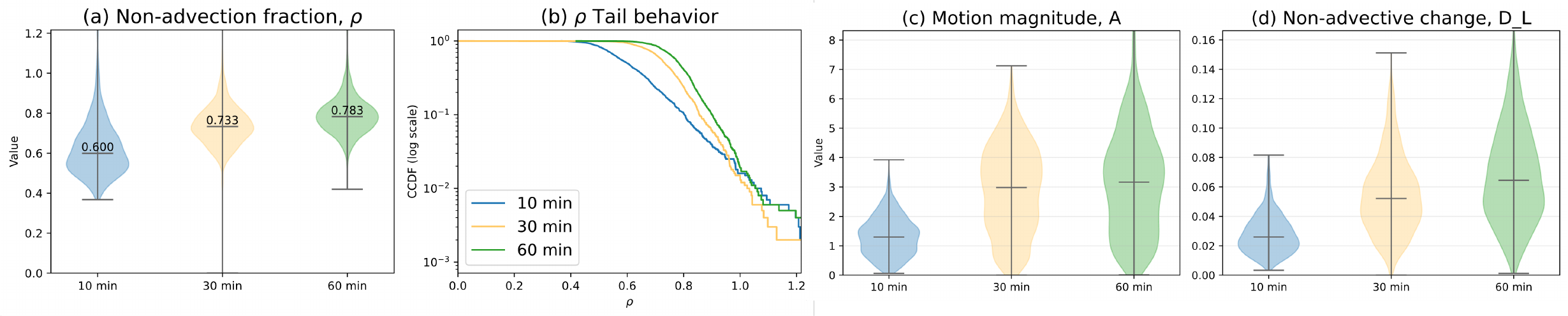}
  \caption{\textbf{Temporal dynamics across RAPID.} As the time interval increases (10-min, 30-min, and 60-min), the \textit{non-advective fraction $\rho$} shifts upward, structural evolution plays a progressively larger role relative to displacement at longer horizons. The broader spread at 30 minutes suggests a transition between motion-dominant and evolution-driven regimes.}
  \label{fig:data-violin}
\end{figure}

\subsection{Dataset Statistics}

\noindent\textbf{Intensity Distribution.}
All statistics are computed in physical units (training split) after inverse decoding. 
In ~\tabref{tab:dataset_stats}, RAPID exhibits a heavier upper tail than SEVIR, particularly in extreme rainfall regimes. 
Although 72\% of pixels correspond to zero rainfall, high-percentile intensities are substantially larger.
Conditioned on non-zero pixels, the 99th percentile reaches 21.28 mm/h in RAPID compared to 13.11 mm/h in SEVIR, indicating stronger separation among intense precipitation events.

To preserve sensitivity across the full dynamic range during training, rainfall intensity is encoded into $[0,255]$ using a piecewise transformation.

\noindent\textbf{Temporal Dynamics Across Intervals.}
RAPID provides forecasting configurations at 10-, 30-, and 60-minute intervals, enabling systematic evaluation of interval-dependent precipitation dynamics.

To quantify how evolution varies with temporal scale, we adopt a motion--evolution decomposition inspired by~\cite{kim2024ace}. 
Optical flow estimates advective displacement, and residual intensity change after warping measures non-advective evolution. 
We report motion magnitude ($A$), non-advective change ($D_L$), total change ($D_E$), and the non-advective fraction $\rho = D_L / (D_E + \epsilon)$.
As shown in ~\figref{fig:data-violin}, the non-advective fraction increases consistently with temporal interval. 
This indicates that rigid displacement explains a decreasing proportion of observed change as lead time extends, while structural evolution becomes increasingly dominant.
By providing temporally distinct configurations with systematically varying dynamic characteristics, RAPID establishes a controlled benchmark for evaluating forecasting models under heterogeneous temporal regimes.
In particular, it enables assessment of models designed to represent precipitation as continuous temporal evolution rather than fixed-interval frame prediction.

\section{Experiments}
\subsection{Experimental Setting}
\noindent\textbf{Dataset.}
We evaluate our method on SEVIR~\cite{veillette2020sevir} and RAPID.  SEVIR is one of the most widely adopted benchmarks for precipitation nowcasting, enabling direct comparison with prior work. 
Other datasets~\cite{shi2017deep,larvor2020meteonet} share similar lead-time configurations with SEVIR, and therefore do not provide additional variation in temporal structure.
Instead, to evaluate forecasting behavior across heterogeneous temporal intervals, we introduce RAPID, which supports 10-, 30-, and 60-minute configurations within a unified benchmark.

\begin{itemize}

\item \textbf{SEVIR} predicts 60 minutes (12 frames) of future VIL from 70 minutes (13 frames) of context at $384 \times 384$ resolution.
We follow the standard temporal split, using samples before June 1, 2019 for training and those after for testing.

\item \textbf{RAPID} evaluates forecasting at 10-, 30-, and 60-minute intervals, corresponding to 1-, 3-, and 6-hour lead times.
Each sample consists of six input and six future frames at $224 \times 224$ resolution.
Data from 2022--2024 are used for training and 2025 for testing.
\end{itemize}

\noindent\textbf{Evaluation Metrics.}
We report both categorical and pixel-wise metrics.  Critical Success Index (CSI) evaluates event-level detection performance. We report mean CSI (CSI-M) and threshold-based CSI (CSI-th) at 16, 160, and 219, including evaluations under $4\times4$ and $16\times16$ pooling to assess robustness to spatial displacement. 
We further report Heidke Skill Score (HSS) and Fractions Skill Score (FSS) for spatially tolerant evaluation, along with RMSE and FAR to measure pixel-wise intensity error and false alarms.  

\input{tables/table_mainresults.tex}

\noindent\textbf{Implementation Details.}
All experiments are conducted on a single NVIDIA H200 GPU (140 GB memory). 
We train all models using the AdamW optimizer with an initial learning rate of $1\times10^{-3}$. 
A cosine annealing scheduler is applied with $T_{\max}=100$ and $\eta_{\min}=1\times10^{-6}$. 
The training objective is the mean squared error (MSE) loss and the weighting coefficient for the consistency loss is set to $\alpha=0.1$.
Models are trained for 100 epochs with a batch size of 12 for RAPID and 4 for SEVIR.
The checkpoint with the highest CSI-M on the validation set is selected for evaluation. The SSM module is implemented strictly following \cite{li2023bbdm}.

All baseline models are trained from scratch in our H200 environment using their official implementations. 
No pretrained weights are used unless explicitly stated. 
3D U-Net, SimVP, and Earthformer are implemented based on the FACL repository\footnote{\url{https://github.com/argenycw/FACL}}, 
while PreDiff\footnote{\url{https://github.com/gaozhihan/PreDiff}}, 
CasCast\footnote{\url{https://github.com/OpenEarthLab/CasCast}}, and 
exPreCast\footnote{\url{https://github.com/tony890048/exPreCast}} strictly follow the official code released by their authors. 
Implicit forecasting models such as MetNet~\cite{sonderby2020metnet} are excluded due to the absence of official implementations and differences in input modalities.

\subsection{Main Results}
\begin{figure}[tb]
  \centering
  \includegraphics[width=1\linewidth]{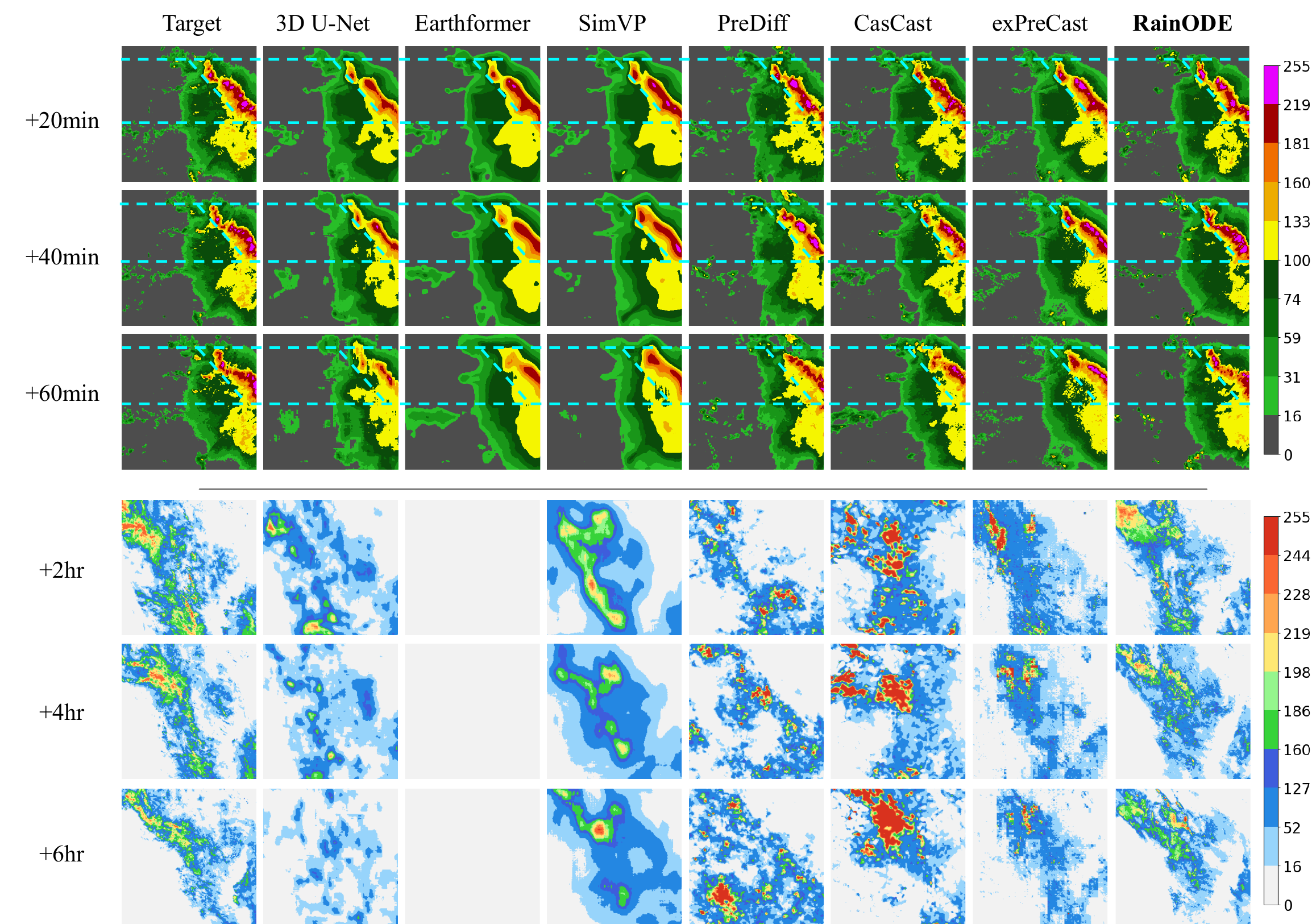}
  \caption{Qualitative results on \textbf{SEVIR and RAPID-60min}. (top) SEVIR forecasts at +20, +40, and +60 minutes. (bottom) RAPID-60min forecasts at +2, +4, and +6 hours. Red dashed lines are overlaid to visually assess spatial alignment of predicted precipitation structures. Pixel values are encoded in [0,255], with higher values corresponding to extreme rainfall.}
  \label{fig:exp_sevir}
\end{figure}  

\noindent\textbf{Quantitative Result.}
Table~\ref{tab:ex-main} presents the quantitative comparison on SEVIR and RAPID (10-, 30-, and 60-minute intervals), evaluating whether RainODE generalizes across both a widely used public benchmark and our proposed multi-interval dataset.
As shown in \tabref{tab:ex-main}, RainODE achieves strong performance across most evaluation metrics on both datasets, consistently outperforming existing spatiotemporal and generative forecasting baselines across multiple temporal intervals.
The performance gap becomes more pronounced on RAPID-60min. 
RainODE maintains stable categorical and spatial skill scores, whereas most discrete-time baselines degrade significantly. 
This result suggests that modeling precipitation dynamics in continuous-time improves robustness under longer forecasting horizons.

Notably, Earthformer exhibits near-zero CSI scores at the 60-minute setting, indicating severe performance degradation under extended lead times.
All methods are trained under identical settings in our H200 environment using official implementations without pretrained weights, ensuring a controlled and reproducible comparison.

\begin{table}[t!]
    \centering
    \caption{CSI-M performance of \textbf{RainODE across lead times}. $T$ denotes the last observed timestep, and $T+1$ corresponds to the first forecasted frame. Performance gradually degrades as the lead time increases.}
    \label{tab:csim_single_model}
    \renewcommand{\arraystretch}{1}
    \setlength{\tabcolsep}{7pt} 
    \begin{tabular}{@{}l| c c c c c c|c@{}}
    \toprule
    \textbf{Dataset} & $T+1$ & $T+2$ & $T+3$ & $T+4$ & $T+5$ & $T+6$ & \textbf{Lead time} \\
    \midrule
    RAPID-10min & 0.489 & 0.459 & 0.430 & 0.405 & 0.382 & 0.360 & 1-hour\\ 
    RAPID-30min & 0.360 & 0.313 & 0.278 & 0.250 & 0.226 & 0.202 & 3-hour\\
    RAPID-60min & 0.298 & 0.249 & 0.221 & 0.200 & 0.170 & 0.144 & 6-hour\\
    \bottomrule
    \end{tabular}
\end{table}
\noindent\textbf{Qualitative Result.}
These trends are also reflected in the qualitative comparisons shown in \figref{fig:exp_sevir}, which presents results on SEVIR (10-minute interval) and RAPID-60min.
On SEVIR, models trained solely with MSE loss (e.g., 3D U-Net, Earthformer, and SimVP) capture the spatial location of precipitation reasonably well, but often underestimate rainfall intensity. 
In contrast, generative or diffusion-based models such as PreDiff and CasCast better preserve intensity patterns, yet exhibit noticeable spatial misalignment.
Our method combines Neural ODE-based trajectory modeling for spatial evolution with SSM for intensity refinement, resulting in improved alignment of both precipitation location and intensity.

The differences become more pronounced on RAPID-60min. Pure MSE-based models (3D U-Net and SimVP) maintain relatively accurate spatial displacement but struggle with intensity calibration.
Earthformer predicts near-uniform low-intensity rainfall across wide regions, leading to visually diminished precipitation structures. PreDiff and CasCast exhibit larger spatial deviations at extended lead times, exPreCast shows visible grid-like artifacts, likely caused by the combination of trilinear interpolation and 3D pixel-shuffle operations within its CDU block.

Although our method provides the most coherent alignment between location and intensity among the compared models, noticeable errors remain. 
Forecasting up to +6 hours from radar-only observations remains inherently challenging, highlighting the intrinsic difficulty of long-horizon precipitation prediction.

\subsection{Continuous-Time Generalization}
\begin{figure}[t!]
  \centering
  \includegraphics[width=1.0\linewidth]{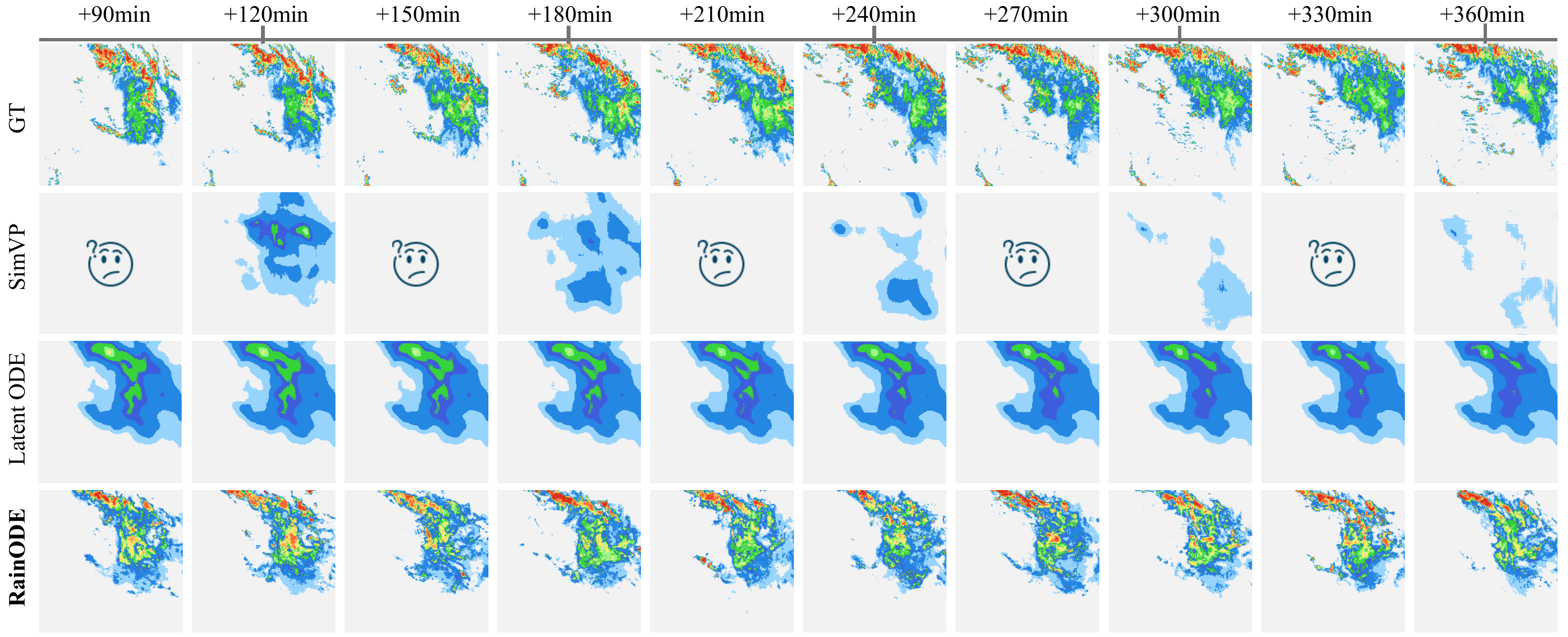}
  \caption{\textbf{Continuous-time forecasting up to 6 hours.} Discrete forecasting models such as SimVP produce predictions only at predefined timesteps and progressively dissipate precipitation intensity over extended horizons. In contrast, Latent ODE (Stage I of RainODE) and the full RainODE model continuous latent dynamics, enabling predictions at arbitrary time points while maintaining coherent precipitation structures.}
  \label{fig:results_300}
\end{figure} 
\noindent\textbf{Temporal Dynamics Across Lead Times.}
Table~\ref{tab:csim_single_model} reports the CSI-M performance of our model across increasing lead times.
Across all temporal configurations, performance consistently decreases as the lead time increases.
This trend is consistent with prior precipitation forecasting studies, where longer lead times inherently introduce greater uncertainty.
As the forecast horizon extends, small discrepancies in the predicted motion and intensity evolution accumulate over time.
Moreover, precipitation systems exhibit increasingly complex structural changes, including growth, decay, merging, and splitting, which are difficult to model accurately over extended periods.

Consequently, the degradation in CSI-M reflects increasing dynamical uncertainty and structural evolution at longer lead times.
Nevertheless, the gradual decline observed in \tabref{tab:csim_single_model} suggests that our model maintains temporal consistency, which we further examine qualitatively.

Figure~\ref{fig:results_300} qualitatively compares three models trained at 1-hour intervals: SimVP, a variant using only the Latent ODE (without the SSM), and the full RainODE.
Since SimVP is trained at fixed 1-hour intervals, it can only produce predictions at those discrete timestamps and does not support intermediate inference. In contrast, RainODE enables continuous-time inference and can generate predictions at arbitrary temporal resolutions between hourly steps.

The comparison highlights a key difference between discrete and continuous-time forecasting. While SimVP gradually dissipates precipitation structures, RainODE maintains coherent spatial trajectories through continuous latent dynamics.
The Latent ODE primarily captures large-scale motion and directional evolution,  preserving the overall displacement of precipitation systems.
The SSM module further refines high-frequency structures and intensity patterns.
As a result, the full RainODE produces forecasts that better preserve precipitation structure and intensity compared to the Latent ODE-only variant.

\begin{figure}[t!]
  \centering
  \includegraphics[width=1.0\linewidth]{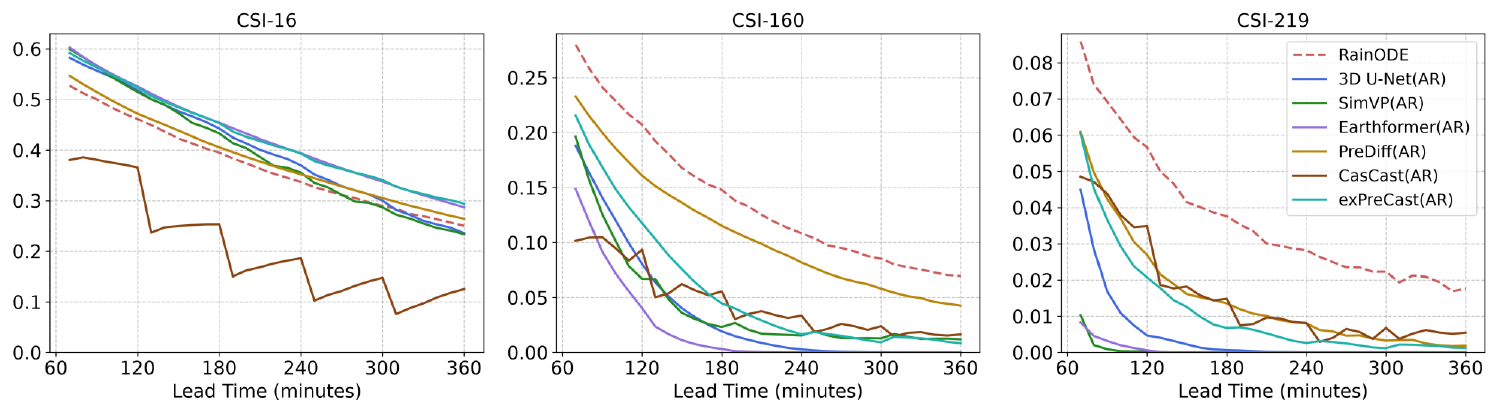}
  \caption{\textbf{The robustness of continuous modeling.} Performance is measured using CSI-16/160/219 at 10-minute evaluation intervals. RainODE is trained at 1-hour intervals and evaluated at finer temporal resolution without retraining, while baseline models rely on autoregressive (AR) rollout for long-horizon prediction.}
  \label{fig:results_lt}
\end{figure}

\noindent\textbf{The Robustness of Continuous Modeling.}
As illustrated in ~\figref{fig:results_lt}, we evaluate whether a model trained at coarse temporal intervals can generalize to finer inference resolutions. 
Specifically, RainODE is trained using 1-hour intervals to predict up to +6 hours, while inference is performed at a 10-minute resolution without any retraining.
Discrete baselines are trained at 10-minute intervals and extended to +6 hours through autoregressive (AR) rollout.

As shown in ~\figref{fig:results_lt}, under the CSI-16 metric, RainODE achieves slightly lower scores compared to AR-based baselines. 
We attribute this behavior to the tendency of AR models to produce smoother and more spatially diffused predictions, which can artificially increase detection scores at low intensity thresholds. 
However, this trend reverses at higher thresholds (CSI-160 and CSI-219), and RainODE outperforms the baselines with a noticeable margin.

Importantly, RainODE is trained only with 1-hour interval data, yet it remains competitive when evaluated at 10-minute resolution. 
This observation suggests that modeling precipitation dynamics in continuous-time represents temporal evolution as a coherent trajectory rather than a sequence of independent predictions.
Compared to autoregressive discrete-time models, which accumulate prediction errors through repeated rollouts, RainODE supports both long-horizon forecasting and dense temporal inference within a single framework.

\begin{figure}[t]
  \centering
  \includegraphics[width=1.0\linewidth]{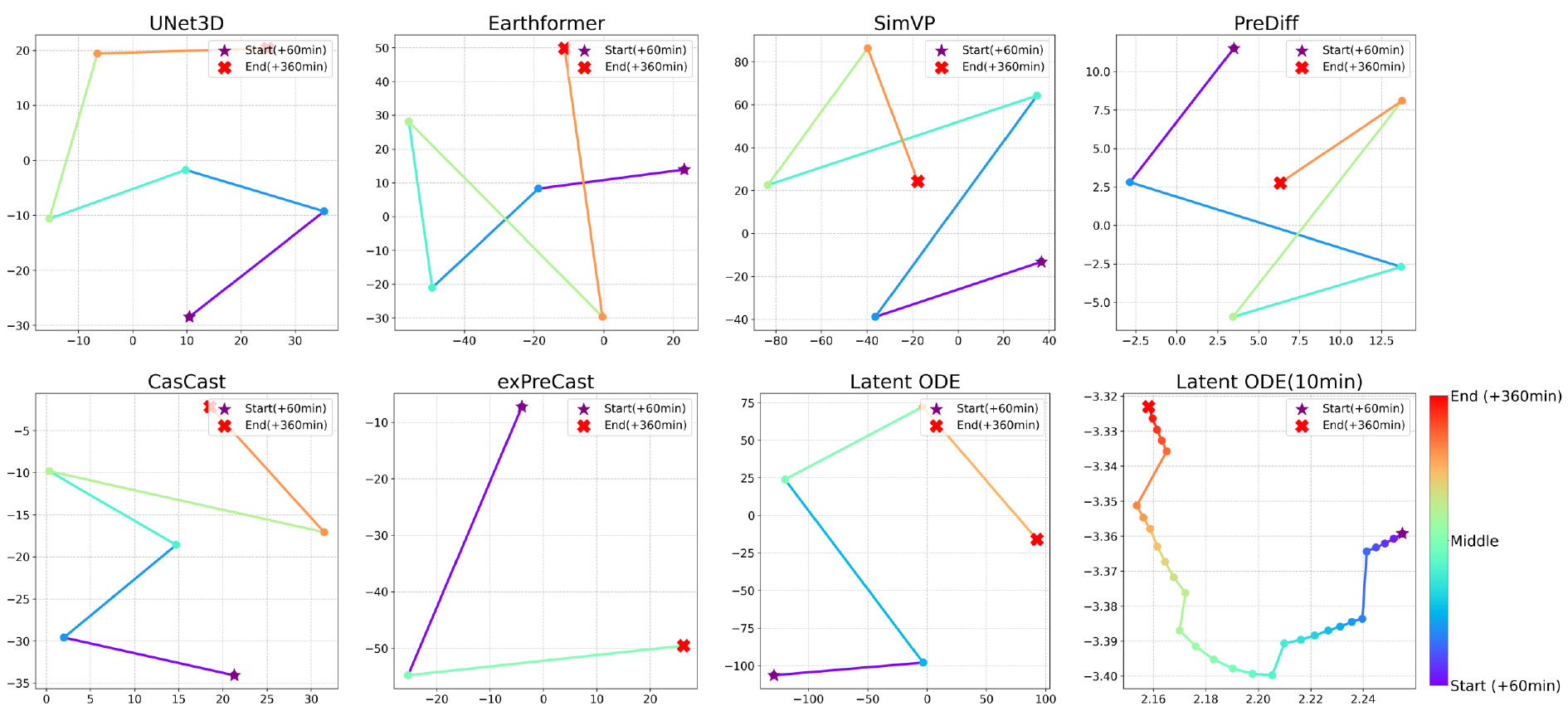}
  \caption{\textbf{Temporal consistency of latent trajectories on RAPID-60min.} Latent ODE (Stage I of RainODE) is additionally evaluated at 10-minute intervals to visualize the continuous-time evolution. Colors indicate temporal progression, with warmer colors approaching the final prediction time (+360 minutes).} 
  \label{fig:tsne}
\end{figure}

\noindent\textbf{Temporal Consistency in Latent Space.} 
To evaluate the long-horizon, temporally dense, and stable forecasting behavior of Latent ODE, we analyze latent-space trajectories. ~\figureref{fig:tsne} presents visualizations of latent representations from the same sample used in ~\figref{fig:exp_sevir}, drawn from the RAPID-60min setting. All latent states from the compared models are jointly embedded using t-SNE to illustrate their temporal progression in the learned representation space.
Among the compared models, Latent ODE shows a more coherent progression from the prediction start point to the final horizon, with fewer abrupt directional changes across forecast steps. In contrast, the latent trajectories of the baseline models are more irregular and often show zigzag patterns, suggesting less coherent temporal progression in latent space. ExPreCast shows a relatively sparse trajectory, which may be attributed to the compression of temporal information in its latent representation.
Beyond the original 1-hour prediction setting, Latent ODE also maintains coherent latent transitions when evaluated at 10-minute intervals without additional training. While these visualizations are qualitative, they indicate that Latent ODE preserves temporal consistency in latent evolution more effectively than discrete-time baselines and supports temporally dense forecasting over extended horizons.

\subsection{Ablations}
Table~\ref{tab:combined_ablation} evaluates the contribution of each stage in RainODE. When integrating SimVP with the Latent ODE alone, CSI-16 and RMSE improve, indicating that continuous deterministic dynamics effectively capture the overall precipitation motion. However, this produces smoother predictions and does not improve CSI-219, suggesting that it is insufficient for concentrated heavy-rainfall cores. By further incorporating SSM, RainODE substantially improves high-intensity detection (CSI-219) and perceptual quality metrics (LPIPS and SSIM), demonstrating its role in restoring high-frequency and intensity-sensitive structures. Although this refinement leads to a small trade-off in RMSE and CSI-16, this is consistent with the role of SSM, which prioritizes sharper and more localized precipitation structures.

From a computational perspective, the proposed framework remains efficient. The Latent ODE model maintains the same number of parameters even when the prediction length increases. Also, FLOPs remain lower than SimVP, although they increase with additional integration steps. In contrast, SimVP requires more parameters as the output horizon expands and generally exhibits higher FLOPs.

\begin{table}[tb]
    \centering
    \caption{Ablation study on the proposed framework. (a) Importance of each stage integrating SimVP, Latent ODE, and SSM(RainODE). (b) Computational cost in parameters (M) and FLOPs (G). $L_{\mathrm{out}}$ denotes the length of output sequence.}
    \label{tab:combined_ablation}
    \renewcommand{\arraystretch}{1.1}
    \setlength{\tabcolsep}{4pt}
    \resizebox{\linewidth}{!}{%
        \begin{tabular}[t]{@{}c | c c c c c c c@{}}
        \multicolumn{8}{c}{(a) Importance of each stage} \\
        \toprule
        \textbf{Method} & \textbf{CSI-M$\uparrow$} & \textbf{CSI-16$\uparrow$} & \textbf{CSI-219$\uparrow$} & \textbf{RMSE$\downarrow$} & \textbf{FSS$\uparrow$} & \textbf{LPIPS$\downarrow$} & \textbf{SSIM$\uparrow$} \\
        \midrule
        SimVP & 0.141 & 0.400 & 0.011 & 0.107 & 0.130 & 0.346 & 0.580  \\
        Latent ODE & 0.152 & \textbf{0.404} & 0.002 & \textbf{0.101} & 0.137 & 0.376 & 0.588 \\
        RainODE & \textbf{0.220} & 0.400 & \textbf{0.056} & 0.113& \textbf{0.223}  & \textbf{0.275} & \textbf{0.690}  \\
        \bottomrule
        \end{tabular}
        
        \quad\quad 
        
        \begin{tabular}[t]{@{}l c c c@{}}
        \multicolumn{4}{c}{(b) Computational cost} \\      
        \toprule
        \textbf{Model} & \textbf{$L_{\mathrm{out}}$} & \textbf{Params} & \textbf{FLOPs} \\
        \midrule
        Latent ODE          & 6  & 9.32  & 68.84 \\
        Latent ODE          & 36  & 9.32  & 104.97 \\
        \midrule
        SimVP    & 6  & 13.62 & 90.91 \\
        SimVP & 36 & 15.25 & 129.78 \\
        \bottomrule
        \end{tabular}
    }
\end{table}

\section{Limitations and Future Work}
We introduced RainODE, which enables inference at unseen temporal intervals through continuous-time modeling. Although it generalizes well across resolutions, our validation is limited to radar-only forecasting.
As shown in the RAPID-60min results, long-term prediction from radar alone has intrinsic limitations. With increasing lead time, precipitation evolution depends more on large-scale atmospheric dynamics that cannot be fully captured from radar observations. 
This fundamentally constrains performance and makes it difficult to predict precipitation that emerges from completely clear conditions.
In future work, we aim to extend RainODE to multi-modal settings, incorporating additional atmospheric variables to improve long-horizon stability and structural consistency.

\section{Conclusion}
We presented RainODE, a continuous-time precipitation forecasting framework that reformulates frame-based prediction as latent trajectory modeling. By parameterizing temporal evolution through a Neural ODE with stochastic refinement, our approach enables flexible and stable forecasting across varying temporal intervals. Experiments on SEVIR and RAPID demonstrate consistent gains at high-intensity thresholds, improved structural coherence, and stronger long-horizon stability compared to discrete recursive methods. These results suggest that continuous-time latent modeling offers a promising paradigm for precipitation forecasting beyond fixed temporal grids.

\bibliographystyle{splncs04}
\bibliography{main}
\end{document}

%% file: tables/table_mainresults.tex
\begin{table*}[tb]
\centering
\caption{Quantitative results on \textbf{SEVIR and RAPID} (10-, 30-, 60-minute intervals).  
All metrics are reported on a 0--1 scale.  
CSI-th is evaluated at intensity thresholds $\{16,160,219\}/255$.  
$\uparrow$ indicates higher is better and $\downarrow$ indicates lower is better.  
The best score in each column is highlighted in  \textcolor{green}{green}.}
\label{tab:ex-main}
\renewcommand{\arraystretch}{1.1}
\setlength{\tabcolsep}{4pt} 
\resizebox{1\columnwidth}{!}{%
\begin{tabular}{ll|cccccc|ccccc} 
\toprule
\multirow{2}{*}{\textbf{Data}} & \multirow{2}{*}{\textbf{Method}} & 
\multicolumn{3}{c}{\textbf{CSI-M}~$\uparrow$} & 
\multicolumn{3}{c|}{\textbf{CSI-th}~$\uparrow$} & 
\multirow{2}{*}{\textbf{RMSE}~$\downarrow$} & 
\multirow{2}{*}{\textbf{FSS}~$\uparrow$} & 
\multirow{2}{*}{\textbf{HSS}~$\uparrow$} & 
\multirow{2}{*}{\textbf{FAR}~$\downarrow$} \\ 

\cmidrule(lr){3-5} \cmidrule(lr){6-8}
& & P1 & P4 & P16 & 16 & 160 & 219 & & & & \\ 
\midrule

\multirow{8}{*}{\rotatebox[origin=c]{90}{\textbf{SEVIR}}} 
& 3D U-Net \textcolor{gray}{\scriptsize (MICCAI '16)}     & 0.332 & 0.334 & 0.361 & 0.746 & 0.188 & 0.038 & 0.064 & 0.461 & 0.431 & \cellcolor{green!20}\textbf{0.238}  \\
& SimVP \textcolor{gray}{\scriptsize (CVPR '22)}        & 0.415 & 0.419 & 0.438 & 0.761 & 0.290 & 0.131 & 0.060 & 0.554 & 0.539 & 0.308 \\
& Earthformer \textcolor{gray}{\scriptsize (NeurIPS '22)}  & 0.426 & 0.425 & 0.435 & \cellcolor{green!20}\textbf{0.768} & 0.300 & 0.134 & \cellcolor{green!20}\textbf{0.057} & \cellcolor{green!20}\textbf{0.565} & 0.550 & 0.284\\
& PreDiff \textcolor{gray}{\scriptsize (NeurIPS '23)}     & 0.338 & 0.365 & 0.454 & 0.699 & 0.200 & 0.086 & 0.078 & 0.486 & 0.448 & 0.505 \\
& CasCast  \textcolor{gray}{\scriptsize (ICML '24)}    & 0.415 & 0.450 & 0.540 & 0.742 & 0.292 & 0.160 & 0.068 & 0.442 & 0.542 & 0.437 \\ 
& exPreCast \textcolor{gray}{\scriptsize (ICLR '26)}   & 0.420 & 0.452 & 0.538 & 0.755 & 0.296 & 0.145 & 0.062 & 0.557 & 0.545 & 0.404\\ 
& Latent ODE & 0.425 & 0.426 & 0.437 & 0.762 & 0.302 & 0.148 & 0.059 & 0.558 & 0.551 & 0.284 \\
& \textbf{RainODE} 
& \cellcolor{green!20}\textbf{0.430} 
& \cellcolor{green!20}\textbf{0.464} 
& \cellcolor{green!20}\textbf{0.544} 
& 0.766 
& \cellcolor{green!20}\textbf{0.304} 
& \cellcolor{green!20}\textbf{0.177} 
& 0.060 
& 0.564 
& \cellcolor{green!20}\textbf{0.558} 
& 0.399 \\
\midrule

\multirow{8}{*}{\rotatebox[origin=c]{90}{\textbf{RAPID-10min}}} 
& 3D U-Net \textcolor{gray}{\scriptsize (MICCAI '16)}  & 0.395 & 0.413 & 0.470 & \cellcolor{green!20}\textbf{0.682} & 0.340 & 0.130 & 0.072 & 0.458 & 0.529 & 0.257  \\
& SimVP  \textcolor{gray}{\scriptsize (CVPR '22)}   & 0.399 & 0.404 & 0.444 & 0.673 & 0.338 & 0.178 & 0.073 & 0.426 & 0.537 & 0.289 \\
& Earthformer \textcolor{gray}{\scriptsize (NeurIPS '22)} & 0.406 & 0.382 & 0.374 & 0.675 & 0.330 & 0.159 & \cellcolor{green!20}\textbf{0.067} & 0.411 & 0.542 & \cellcolor{green!20}\textbf{0.224} \\
& PreDiff \textcolor{gray}{\scriptsize (NeurIPS '23)}  & 0.375 & 0.425 & 0.521 & 0.621 & 0.337 & 0.140 & 0.084 & 0.449 & 0.509 & 0.411\\
& CasCast \textcolor{gray}{\scriptsize (ICML '24)} & 0.399 & 0.470 & 0.585 & 0.640 & 0.384 & 0.137 & 0.084 & 0.509 & 0.534 & 0.462 \\ 
& exPreCast \textcolor{gray}{\scriptsize (ICLR '26)} & 0.389 & 0.455 & 0.568 & 0.661 & 0.332 & 0.165 & 0.076 & 0.483 & 0.526 & 0.334 \\ 
& Latent ODE & \cellcolor{green!20}\textbf{0.443} & 0.427 & 0.417 & 0.676 & 0.399 & \cellcolor{green!20}\textbf{0.197} & \cellcolor{green!20}\textbf{0.067} & 0.451 & \cellcolor{green!20}\textbf{0.583} & 0.279 \\
& \textbf{RainODE} 
& 0.420
& \cellcolor{green!20}\textbf{0.491} 
& \cellcolor{green!20}\textbf{0.597} 
& 0.618 
& \cellcolor{green!20}\textbf{0.416} 
& 0.187
& 0.082 
& \cellcolor{green!20}\textbf{0.514} 
& 0.561
& 0.418 \\
\midrule

\multirow{8}{*}{\rotatebox[origin=c]{90}{\textbf{RAPID-30min}}} 
& 3D U-Net \textcolor{gray}{\scriptsize (MICCAI '16)}  & 0.197 & 0.216 & 0.280 & 0.485 & 0.124 & 0.035 & 0.099 & 0.185 & 0.285 & 0.469 \\
& SimVP   \textcolor{gray}{\scriptsize (CVPR '22)} & 0.224 & 0.225 & 0.238 & 0.512 & 0.154 & 0.026 & 0.101 & 0.196 & 0.320 & 0.458 \\
& Earthformer \textcolor{gray}{\scriptsize (NeurIPS '22)} & 0.161 & 0.165 & 0.165 & 0.462 & 0.040 & 0.000 & 0.097 & 0.109 & 0.221 & 0.452 \\
& PreDiff \textcolor{gray}{\scriptsize (NeurIPS '23)} & 0.138 & 0.180 & 0.305 & 0.339 & 0.091 & 0.023 & 0.128 & 0.135 & 0.200 & 0.702 \\
& CasCast \textcolor{gray}{\scriptsize (ICML '24)} & 0.234 & 0.290 & 0.396 & 0.445 & 0.215 & 0.043 & 0.123 & 0.234 & 0.334 & 0.671 \\ 
& exPreCast \textcolor{gray}{\scriptsize (ICLR '26)} & 0.234 & 0.290 & 0.437 & 0.485 & 0.174 & 0.055 & 0.101 & 0.267 & 0.336 & 0.561 \\ 
& Latent ODE & 0.234 & 0.237 & 0.256 & \cellcolor{green!20}\textbf{0.517} & 0.166 & 0.026 & \cellcolor{green!20}\textbf{0.093} & 0.206 & 0.333 & \cellcolor{green!20}\textbf{0.429} \\
& \textbf{RainODE} 
& \cellcolor{green!20}\textbf{0.273} 
& \cellcolor{green!20}\textbf{0.340} 
& \cellcolor{green!20}\textbf{0.458} 
& 0.491 
& \cellcolor{green!20}\textbf{0.244} 
& \cellcolor{green!20}\textbf{0.080} 
& 0.105 
& \cellcolor{green!20}\textbf{0.288} 
& \cellcolor{green!20}\textbf{0.391} 
& 0.556 \\
\midrule

\multirow{8}{*}{\rotatebox[origin=c]{90}{\textbf{RAPID-60min}}} 
& 3D U-Net \textcolor{gray}{\scriptsize (MICCAI '16)} & 0.132 & 0.146 & 0.195 & 0.351 & 0.080 & 0.022 & 0.109 & 0.131 & 0.197 & 0.635 \\
& SimVP \textcolor{gray}{\scriptsize (CVPR '22)}   & 0.141 & 0.145 & 0.155 & 0.400 & 0.061 & 0.011 & 0.107 & 0.130 & 0.203 & 0.626 \\
& Earthformer \textcolor{gray}{\scriptsize (NeurIPS '22)} & 0.000 & 0.000 & 0.000 & 0.000 & 0.000 & 0.000 & 0.118 & 0.068 & 0.000 & 0.000 \\
& PreDiff \textcolor{gray}{\scriptsize (NeurIPS '23)}  & 0.064 & 0.085 & 0.144 & 0.177 & 0.036 & 0.009 & 0.121 & 0.092 & 0.099 & 0.729 \\
& CasCast \textcolor{gray}{\scriptsize (ICML '24)} & 0.033 & 0.053 & 0.112 & 0.107 & 0.017 & 0.003 & 0.228 & 0.035 & 0.006 & 0.953 \\ 
& exPreCast \textcolor{gray}{\scriptsize (ICLR '26)}  & 0.140 & 0.172 & 0.305 & 0.363 & 0.079 & 0.021 & 0.112 & 0.128 & 0.205 & 0.710 \\
& Latent ODE & 0.152 & 0.155 & 0.156 & \cellcolor{green!20}\textbf{0.404} & 0.072 & 0.002 & \cellcolor{green!20}\textbf{0.101} & 0.137 & 0.217 & \cellcolor{green!20}\textbf{0.560} \\
& \textbf{RainODE}
& \cellcolor{green!20}\textbf{0.220} 
& \cellcolor{green!20}\textbf{0.270} 
& \cellcolor{green!20}\textbf{0.358} 
& 0.400 
& \cellcolor{green!20}\textbf{0.201} 
& \cellcolor{green!20}\textbf{0.056} 
& 0.113  
& \cellcolor{green!20}\textbf{0.223} 
& \cellcolor{green!20}\textbf{0.328} 
& 0.596  \\
\bottomrule
\end{tabular}
}
\end{table*}